\documentclass[11pt,a4paper]{article}
\usepackage[hyperref]{eacl2021}
\usepackage{times}
\usepackage{latexsym}

\usepackage{graphicx}

\usepackage{microtype}

\aclfinalcopy % Uncomment this line for the final submission
\title{indicnlp@kgp at DravidianLangTech-EACL2021: Offensive Language Identification in Dravidian Languages}

\author{Kushal Kedia \\
  IIT Kharagpur \\
  \texttt{kushal.k@iitkgp.ac.in} \\
  \And
  Abhilash Nandy \\
  IIT Kharagpur \\
  \texttt{nandyabhilash@iitkgp.ac.in} \\}

\date{}

\begin{document}
\maketitle
\begin{abstract}
The paper presents the submission of the team indicnlp@kgp to the EACL 2021 shared task “Offensive Language Identification in Dravidian Languages”. The task aimed to classify different offensive content types in 3 code-mixed Dravidian language datasets. The work leverages existing state of the art approaches in text classification by incorporating additional data and transfer learning on pre-trained models. Our final submission is an ensemble of an AWD-LSTM based model along with 2 different transformer model architectures based on BERT and RoBERTa. We achieved weighted-average F1 scores of $0.97$, $0.77$, and $0.72$ in the Malayalam-English, Tamil-English, and Kannada-English datasets ranking ${1}^{st}$, ${2}^{nd}$, and $3^{rd}$ on the respective tasks.
\end{abstract}

\section{Introduction}

Offensive language identification is a natural language processing (NLP) text classification task where the goal is to moderate and reduce objectionable social media content. There has been a rapid growth in offensive content on social media and the number of users from different ethnicities and cultures worldwide. A significant portion of offensive content is specifically targeted at various individuals and minority \& ethnic groups. Consequently, the identification and classification of these different kinds of foul language are receiving increased importance. Dravidian languages like Kannada, Malayalam, and Tamil \cite{nui_2020} are low-resourced making this task challenging. Training embeddings of words has previously been a common approach employed in text classification tasks. However, transfer learning approaches in deep learning \cite{mou-etal-2016-transferable} have been shown unsuccessful or requiring extremely large collections of in-domain documents to produce strong results \cite{NIPS2015_7137debd}. 

Further, in a multilingual culture, code-mixing is a prevalent practice and code-mixed texts are sometimes written in native scripts. Due to the complexity introduced by code-switching at multiple linguistic levels, systems trained on monolingual data can fail on code-mixed data. While multilingual versions of transformer models have been shown to perform remarkably well, even in zero-shot settings \cite{pires-etal-2019-multilingual}, a zero-shot transfer may perform poorly or fail altogether \cite{sogaard-etal-2018-limitations}. This is when the target language, here code-mixed Dravidian data, is different from the source language, mainly monolingual.  In our work, we tackle these problems by exploiting additional datasets for fine-tuning our models and using effective transfer learning techniques. Our code and experiments are available on GitHub\footnote{\url{https://github.com/kushal2000/Dravidian-Offensive-Language-Identification}} for generating reproducible results on our models.
% We also open-source our code on GitHub\footnote{\url{https://github.com/kushal2000/Dravidian-Offensive-Language-Identification}} for reproducing the methods applied in our work.

\section{Task Description and Datasets}

The purpose of this task is to classify offensive language material gathered from social media from the a set of code-mixed posts in Dravidian Languages. The systems have to classify each post into one of the 6 labels: 
\begin{itemize}
    \item not offensive
    \item untargeted offense
    \item offense targeted at an individual
    \item offense targeted at a group 
    \item offense targeted at someone else
    \item not in intended language
\end{itemize}
There is also a significant class imbalance in all the datasets representing a real-world situation. This shared task presents a new gold standard corpus for offensive language identification of code-mixed text in three Dravidian languages: Tamil-English \cite{chakravarthi-etal-2020-corpus}, Malayalam-English \cite{chakravarthi-etal-2020-sentiment}, and Kannada-English \cite{hande-etal-2020-kancmd}. The Malayam dataset does not contain the \textit{offense targeted at someone else} tag. The posts can contain more than one sentence, but the average number of sentences is $1$. The Tamil and Malayalam datasets are considerably large containing over $30k$ and $20k$ annotated comments while the Kannada dataset is relatively smaller with almost $8k$ annotations. Apart from the dataset supplied by the organizers, we also use a monolingual English Offensive Language Identification Dataset (OLID) \cite{zampieri-etal-2019-predicting} used in the SemEval-2019 Task 6 (OffensEval) \cite{zampieri-etal-2019-semeval}. The dataset contains the same labels as our task datasets with the exception of the \textit{not in intended language} label. The one-to-one mapping between the labels in OLID and it's large size of $14k$ tweets makes it suitable for aiding the transfer learning detailed in Section~\ref{section:TransformerModels}.

\section{Methods}

A variety of methods are experimented on the datasets to provide a complete baseline. In Section~\ref{section:ML}, we describe our implementation of three traditional machine learning classifiers; Multinomial Naive Bayes, Linear Support Vector Machines (SVM) and Random Forests. These approaches work well on small datasets and are more computationally efficient than deep neural networks. Their performance is similar to the models described in the latter sections in the absence of pretraining and additional data.  In Section~\ref{section:DL}, our Recurrent Neural Network (RNN) models are explained. We have compared an LSTM model using word-level embeddings trained from scratch, to an ULMFiT model, an effective transfer learning approach for language models. Finally, in Section~\ref{section:TransformerModels}, we discuss transformer architectures using their cross-lingual pretrained models which can be data intensive during fine-tuning but provide the strongest results on our datasets.

\subsection{Machine Learning Classifiers} \label{section:ML}

\textbf{Dataset Preprocessing} The datasets are preprocessed by removing punctuation, removing English stop words, removing emojis, and lemmatizing the English Words. The Natural Language Toolkit library \cite{bird-loper-2004-nltk} was used for lemmatization and removing stop words. A word vocabulary is made and vocabulary-length vectors containing counts of each word are used to represent each individual input. Based on the Mutual Information scores of each word, feature selection is done to reduce the vocabulary size. 

\textbf{Hyperparameters}  For all three models, the number of words selected using the top Mutual Information scores was varied from $1000$ to the length of the vocabulary. Further hyperparameters were specific to the SVM and Random Forest. The random state, the regularisation parameters, and max iterations were tuned for the SVM and the number of decision trees used was the only hyperparameter in the case of random forests. 

\subsection{RNN Models} \label{section:DL}

\textbf{Vanilla LSTM} To set a baseline for an RNN approach, we build word embeddings from scratch using just the individual datasets. For this, we selected the top $32,000$ occurring words in each dataset for one-hot encoding, which is passed through an embedding layer to form $100$-dimension word vectors. A spatial dropout of $0.2$ followed by a single LSTM cell and a final softmax activation forms the rest of the model. While the results for larger datasets are marginally better than the previous section, they are worse in comparison to the transfer learning approach.

\textbf{ULMFiT} Transfer learning has been shown to perform well in text classification tasks. Usually, language models are trained on large corpora and their first layer, i.e, the word embeddings are fine-tuned on specific tasks. This approach has been a very successful deep learning approach in many state of the art models. \cite{NIPS2013_9aa42b31} However, \citealp{howard-ruder-2018-universal} argue that we should be able to do better than randomly initializing the remaining parameters of our models and propose \textit{ULMFiT: Universal Language Model Fine-tuning for Text Classification}. For the Dravidian languages in this task, the problem of in-domain data collection for effective transfer is also significant especially in the domains of hate speech. ULMFiT provides a robust framework for building language models from moderate corpora and fine-tunes them on our specific tasks.

\textbf{Language Models \& Corpora} We make use of language models open-soruced by the team \textit{gauravarora} \cite{arora2020gauravarorahasocdravidiancodemixfire2020} in the shared task at HASOC-Dravidian-CodeMix FIRE-2020 \cite{fire2020}. They build their corpora for language modelling from large sets of Wikipedia articles. For Tamil \& Malayalam languages, they also generate code-mixed corpora by obtaining parallel sets of native, transliterated and translated articles and sampling sentences using a Markov process, which has transition probabilities to $3$ states; native, translated and transliterated. For Kannada, only a native script corpus is available and we had to transliterate our code-mixed dataset to Kannada to match their language model. The models are based on the Fastai \cite{Howard_2020} implementation of ULMFiT. Pre-trained tokenizers and language models are available on Github. \footnote{\url{github.com/goru001/nlp-for-tanglish}} \footnote{\url{github.com/goru001/nlp-for-manglish}} \footnote{\url{github.com/goru001/nlp-for-kannada}}

\textbf{Preprocessing \& Model Details} Basic preprocessing steps included lower-casing, removing punctuations and mentions, etc. Subword tokenization using unigram segmentation is implemented which is reasonably resilient to variations in script and spelling.  The tokenization model used is SentencePiece\footnote{\url{github.com/google/sentencepiece}}. The language model is based on an AWD-LSTM \cite{merity2018regularizing}, a regular LSTM cell with additional parameters related to dropout within the cell. The text classification model additionally uses two linear layers followed by a softmax on top of the language model. To tackle the difference in distributions of the target datasets and the pretraining corpora, ULMFiT proposes using 1) \textit{discriminative fine-tuning}, i.e, layers closer to the last layer have higher learning rates, 2) \textit{slanted triangular learning rates} which increase aggressively during the start of training and then decay gradually and 3) \textit{gradual unfreezing}, i.e, instead of learning all layers of the model at once, they are gradually unfrozen starting from the last layer. The combination of these techniques leads to robust transfer learning on our datasets.

\subsection{Transformer Models} \label{section:TransformerModels}
In recent years, transformer networks like the Bidirectional Encoder Representation from Transformer (BERT) \cite{devlin-etal-2019-bert} and its variant RoBERTa \cite{liu2019roberta} have been used successfully in many offensive language identification tasks. For our work, we use the already pre-trained cross-lingual versions of these models available in the HuggingFace\footnote{\url{https://huggingface.co/}} library. Specifically, we use the \textit{bert-base-multilingual-cased} model, mBERT trained on cased text in $104$ languages from large Wikipedia articles, and the \textit{xlm-roberta-base} model, XLM-R \cite{xlmr2020} trained on $100$ languages, using more than two terabytes of filtered CommonCrawl data. Both of these models were originally trained on a masked language modelling objective and we fine-tune them on our specific downstream text classification tasks.

\textbf{Transfer Learning} The core principle of the transfer learning approach is to use a pretrained transformer model for training a classification model on a resource-rich language first, usually English, and transfer the model parameters on a less resource-rich language. For this approach, we concatenate all 3 code-mixed datasets as well as the OLID dataset. Our results do not change significantly on transliteration of all datasets to Roman script. The \textit{not in intended language} label is also removed for fine-tuning on the combined dataset since this label does not represent the same meaning across the datasets. We then use these learned model weights replacing the final linear layer to include the additonal removed label \textit{not in intended language}. This kind of fine-tuning approach has been shown to increase the performance of various scarce-resourced languages such as Hindi and Bengali, etc. \cite{ranasinghe-zampieri-2020-multilingual}.
\begin{figure}[!t]
    \centering
    \includegraphics[width = 8cm]{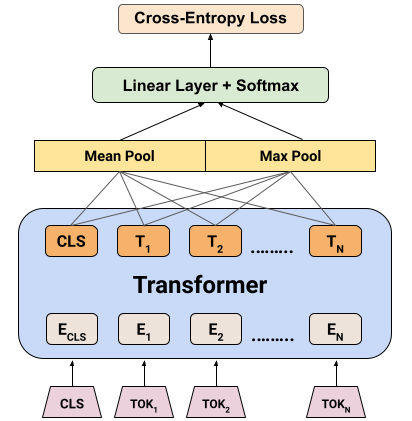}
    \caption{Transformer Model Architecture}
    \label{Transformer}
\end{figure}

\textbf{Model Architecture} We restrict the maximum length of the input sentences to be $256$ by truncation and zero-padding. As shown in Fig~\ref{Transformer}, using the contextual embeddings from the last hidden states of all tokens in the sentence, we build a vector representation by concatenating the max pooling and mean pooling of these hidden states.  Correspondingly, the dimension of the final sentence representation is $1536$ x $1$ . This is passed through a linear layer with a dropout of $0.3$. The learning rate for all fine-tuning was fixed as $2e^{-5}$ and batch size was $32$. The only preprocessing step before feeding the input to the transformer tokenizers was the replacement of emojis by their description in English. This is done because the tokenizers might not recognize the emojis, but they contain useful information about the sentiment of the sentence.

\section{Experiments and Results}
The assessment metric for this task is Weighted-F1, which is the F1 score weighted by number of samples in all the classes. The datasets are divided into train, validation, and test sets in an approximately 8:1:1 ratio. The test labels are hidden and only available to us after the evaluation phase is over. We strictly train our models on the train set using the scores on the validation set for hyperparameter tuning. We have reported the results of our various models on the validation set. The average-ensemble of our top three performing models is submitted finally and we also report it's scores on the validation and test set - using the scores we are ranked with in the task leader board.

\begin{table}[h]
\centering
\begin{tabular}{c |c |c |c } 
\hline \textbf{Model} & \textbf{T} & \textbf{M} & \textbf{K} \\ \hline
Random Forest & 0.69 & 0.94 & 0.62 \\
Naive Bayes & 0.74 & 0.94 & 0.64\\
Linear SVM & 0.74 & 0.95 & 0.65 \\
\hline
\end{tabular}
\caption{\label{ML results} Weighted-F1 scores for ML models on Tamil (T), Malayalam (M) and Kannada (K) datasets.}
\end{table}

Table~\ref{ML results} showcases the scores we have obtained for standard machine learning algorithms. Out of the three traditional machine learning algorithms, the Linear SVM model is best across all three datasets. The results in Table~\ref{DL results} summarize our RNN approaches where ULMFiT is markedly superior. The performance of our transformer models detailed in Table~\ref{Transformer results} considers two settings, one without transfer learning and one with transfer learning using the OLID and other Dravidian code-mixed datasets in conjunction.

\begin{table}[h]
\centering
\begin{tabular}{c |c |c |c } 
\hline \textbf{Model} & \textbf{T} & \textbf{M} & \textbf{K} \\ \hline
Vanilla LSTM & 0.74 & 0.95 & 0.64\\
ULMFiT & 0.76 & 0.96 & 0.71 \\
\hline
\end{tabular}
\caption{\label{DL results} Weighted-F1 scores for RNN models on Tamil (T), Malayalam (M) and Kannada (K) datasets.}
\end{table}

\begin{table}[h]
\centering
\begin{tabular}{c |c |c |c } 
\hline \textbf{Model} & \textbf{T} & \textbf{M} & \textbf{K} \\ \hline
mBERT & 0.74 & 0.95 & 0.66 \\
XLM-R & 0.76 & 0.96 & 0.67\\
mBERT (TL) & 0.75 & 0.97 & 0.71 \\
XLM-R (TL) & 0.78 & 0.97 & 0.72 \\
\hline
\end{tabular}
\caption{\label{Transformer results} Weighted-F1 scores for transformers on Tamil (T), Malayalam (M) and Kannada (K) datasets. TL indicates transfer learning using OLID and other datasets.}
\end{table}
The results on the validation set of our transfer-learnt XLM-R model are the best across all $3$ datasets and is followed closely by the transfer learnt multilingual BERT model and the ULMFiT model. We finally submit an average ensemble of these three models and our results on the validation set and the test set used in the final task evaluation are also enlisted in Table~\ref{Final results} below.

\begin{table}[h]
\centering
\begin{tabular}{c |c |c |c } 
\hline \textbf{Model} & \textbf{T} & \textbf{M} & \textbf{K} \\ \hline
avg-Ensemble (V) & 0.78 & 0.97 & 0.73 \\
avg-Ensemble (T) & 0.77 & 0.97 & 0.72\\
\hline
\end{tabular}
\caption{\label{Final results} Final weighted-F1 scores using average ensembling on Tamil (T), Malayalam (M) and Kannada (K) validation (V) and test (T) datasets.}
\end{table}

\section{Conclusion}

This paper describes various approaches for offensive language identification in three code-mixed English-Dravidian language datasets. We also discuss the final system submitted by the indicnlp@kgp team, which ranks first, second, and third on the competition's three tasks. The benefit of pre-trained language models was shown by the significant improvement in results using a robust transfer learning framework (ULMFiT) compared to a vanilla LSTM model trained from scratch. Transformer networks' performance also improved when all the Dravidian language datasets were combined. This suggests that learning from one Dravidian language may help in zero-shot or few-shot transfer to other new Dravidian languages. In future works, we wish to explore these effects in more detail.

\bibliography{anthology,eacl2021}
\bibliographystyle{acl_natbib}

\end{document}